# Uncertainty in Ontologies: Dempster-Shafer Theory for Data Fusion Applications

Amandine Bellenger[1], Sylvain Gatepaille[1]

[1]EADS, Defence and Security
Information Processing, Control and Cognition department
Parc d'Affaires des Portes - 27106 Val-de-Reuil – France
{amandine.bellenger, sylvain.gatepaille}@eads.com

***Abstract*—Nowadays ontologies present a growing interest in Data Fusion applications. As a matter of fact, the ontologies are seen as a semantic tool for describing and reasoning about sensor data, objects, relations and general domain theories. In addition, uncertainty is perhaps one of the most important characteristics of the data and information handled by Data Fusion. However, the fundamental nature of ontologies implies that ontologies describe only asserted and veracious facts of the world. Different probabilistic, fuzzy and evidential approaches already exist to fill this gap; this paper recaps the most popular tools. However none of the tools meets exactly our purposes. Therefore, we constructed a Dempster-Shafer ontology that can be imported into any specific domain ontology and that enables us to instantiate it in an uncertain manner. We also developed a Java application that enables reasoning about these uncertain ontological instances.**

## I. INTRODUCTION

An ontology contains the concepts used to describe and represent an area of knowledge. By clearly defining semantically a group of terms in the given domain and the relationships among them, without any ambiguity, an ontology encodes the knowledge of the domain in such a way that it can be understood by a computer.

For this to be possible, ontologies must be formalized in languages processable by computers. Without any doubt, the most popular is OWL: Web Ontology Language, which is the latest recommendation of W3C (World Wide Web Consortium), made in February 2004 for the first version 1.0 and more recently at the end of October 2009 for version 2.0. OWL 2 adds new functionalities, which some are said to be only "syntactic sugar" while others really offer new expressivity.

There has been since 2003 a growing interest of ontology-based technologies within data fusion, as shown by the increasing number of papers in international Conferences. For instance during the 6th International Conference on Information Fusion, held in Australia in 2003, was organized a Special Session titled "Ontology Needs and Issues for Higher-Level Fusion."

Data Fusion embodies the process of data combination to refine state estimates and predictions [1]. These data are obtained from various and heterogeneous sensors such as radar, satellite, human messages, etc. The JDL model, elaborated by the Joint Directors of Laboratories Data Fusion Group, presents a pedagogically common understanding of Data Fusion, illustrated on the figure 1 below.

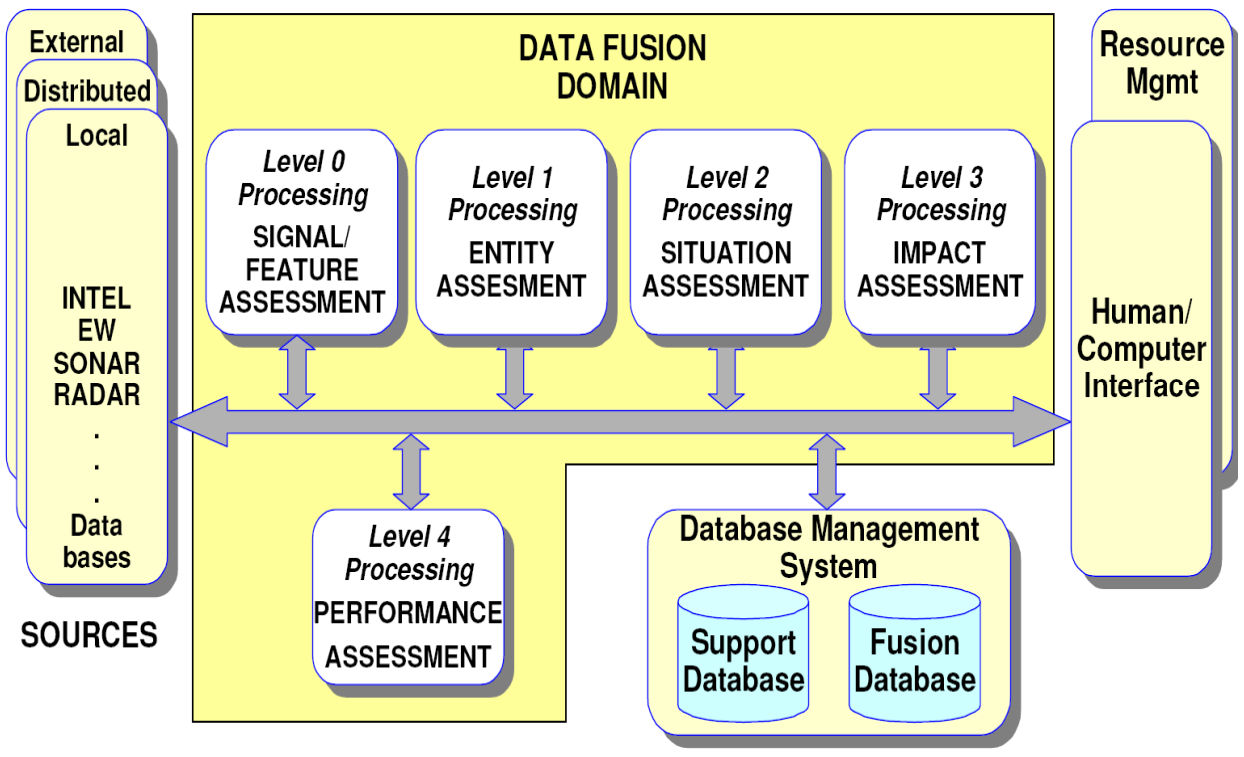

Figure 1. JDL Recommended Revised Data Fusion Model (version from 1998)

In the field of National Defense, community focuses on problems involving the location, characterization, and identification of distinct and dynamic entities such as emitters, platforms, weapons, and military units (level 1). Beyond this task, one seeks higher-level inferences about the enemy situation (such as the relationship among entities, their relationships with the environment and higher level enemy organizations for example, which refers to level 2) and the possible threat they embody (level 3).

In that context, ontologies are useful in providing means for describing and reasoning about sensor data, objects, relations and general domain theories. In that way, they enable to explicitly encode a shared understanding of some domain, in order to disambiguate terms associated with that domain.

However, uncertainty is an important characteristic of our data and information handled by Data Fusion applications. At a low level of the JDL model, observations received from sensors are sullied with uncertainty, due to operational conditions of observations (concealment, jamming, dissemination, etc.). In this paper, we could note that the term "uncertainty" is intended to cover a variety of forms of incomplete knowledge, including incompleteness, vagueness, ambiguity, and others. Uncertainty can be due to measuring errors made by physical sensors (human or electronic). For example, a radar can observe the speed of a vehicle with an error of + or – 3 Km/h. At level 1, it should be necessary to enable the representation of knowledge such as: “Possibility of existence of a regiment position (x,y,t)”. There is here a doubt concerning the existence of the observed entity. At level 2, there could be uncertain relations between entities. For instance, “do two military aircrafts maintain a communication?”. Finally, uncertainty is also ubiquitous in higher levels where the assessment of impact (including threat) is always associated to a certain degree of confidence.

As previously shown, uncertainty is inherent to data. However, uncertainty also naturally occurs in the fusion process itself. Issues related to uncertainty arise in case the set of information acquired from multiple sources about the same fact is inconsistent, or - more generally - in case that multiple information sources attribute different grades of belief to the same statement. If the user/application is not able to decide in favor of a single alternative (due to insufficient trust in the respective information sources), the aggregated statement resulting from the fusion of multiple statements is typically uncertain. The result needs to reflect and weight the different information inputs appropriately, which typically leads to uncertainty.

This paper is organized as follows. In the second section it begins to establish the limitation of traditional ontology formalism in the field of uncertainty representation. Then, it discusses the various ways that have been chosen to represent uncertainty within ontologies, by making a state of art of the most known approaches and associated tools.

In section 3, we will briefly remind what make the Belief Theory a so appealing theory to represent uncertainty within our Data Fusion Applications. Consequently, we will present the ontology that we have built, called DS-Ontology, according to the Dempster-Shafer model, which helps to construct uncertain aspects into any specific domain ontologies. The section will then present the Java application we developed in order to reason formally on an ontology that imports the DS-Ontology, thanks to Jena [5], a Java framework for building Semantic Web applications.

## II. EXISTING APPROACHES FOR UNCERTAINTY REPRESENTATION WITHIN ONTOLOGIES

We will focus here naturally on approaches that directly deal with Semantic languages, and especially OWL. Moreover, we will present some open source tools that we have been able to find out and that permit in some way to represent uncertainty in ontology. We will organize this survey regarding their underlying theory. But first of all, let’s remind why traditional ontologies cannot, on their own, answer this uncertainty issue.

### *A. Limit of traditional ontologies for uncertainty representation*

One of the major limitations of traditional ontology formalisms is the lack of consistent support for uncertainty, reliability and imprecision. As a matter of fact, ontologies currently designed contain only concepts and relations that describe asserted facts about the world. This means that the ontology itself is not uncertain in nature, but rather presents an a priori model of the world, which has to be taken as true by its users. However, in many cases, it is preferable to store a piece of information even imprecise and uncertain rather than to interpret its contents in a restrictive manner, which will lead to store erroneous pieces of information.

There is currently no standard mechanism for interoperable description and exchange of uncertain information over ontologies and even less a standard to reason on this type of information.

However, many researchers are currently trying to enhance ontological capabilities concerning this field. As a matter of fact, the requirement for representation has led the World Wide Web Consortium (W3C) to set up the Uncertainty Reasoning for the World Wide Web Incubator Group in 2007 [2]. It delivered its final report in March 2008, which exposed mainly the different reasoning challenges and the various use cases for the need of uncertainty representation in the Semantic Web.

### *B. Probabilistic Approaches*

Probability domain is surely the most known mathematical theory dealing with uncertainty. In brief, it provides a mathematically sound representation language and formal calculus for rational degrees of belief.

Firstly, some existing tools are based on this theory and more specifically on the Bayesian networks. Bayesian networks (BN) are a powerful probabilistic graphical model to represent a set of random variables and their conditional independencies. Bayesian networks are directed acyclic graphs. The arcs point the direction from the cause to the consequence. BayesOWL [6] is one proposal to model uncertainty in OWL ontologies through BN. It is a mechanism to express OWL ontologies as BN by adding a second ontology to this translation which declares the probabilistic relationships. The advantage is that neither OWL nor ontologies defined in OWL need to be modified. It is used to quantify the degree of the overlap or inclusion between two concepts. However, there are important limitations considering the translation of ontologies. As a matter of fact, it does not take into account neither the properties represented by OWL, the instances nor the specific datatypes.

[2] further details on: http://www.w3.org/2005/Incubator/urw3/

Secondly, there are also the so-called probabilistic extensions to Description Logics (DL) that want to be seen as an alternative to more traditional Bayesian approaches. For instance, Pronto [7] is a probabilistic DL reasoner prototype. Pronto is able to represent and reason about uncertainty in both, generic background knowledge and individual facts (respectively probabilistic relationships between OWL classes and relationships between an OWL class and an individual). We could note that Pronto uses OWL version 1.1 axiom annotations (which is however not a W3C recommendation) to associate probability intervals with uncertain OWL axioms. The main advantages are clearly the ease of the representation and of the reasoner manipulation. However this tool presents really serious deficiencies, considering its scalability and speed of processing. Most notably, Pronto's performance does not scale past beyond around 15 generic conditional constraints in an OWL knowledge base.

Finally, there are also probabilistic approaches that rely on first-order logic such as PR-OWL [8] in combination with UnBBayes-MEBN. PR-OWL is an upper ontology. In order to use it, one has to import the PR-OWL ontology into an ontology editor (e.g. Protégé). Then one can start constructing our domain-specific concepts using the PR-OWL definitions to represent uncertainty about their attributes and relationships according to the Multi-Entity Bayesian Network (MEBN) model. A graphical user interface (UnBBayes-MEBN) is also available to make it easier to instantiate the main ontological concepts (M-Frags, Resident, Input, Context Nodes, etc). UnBBayes-MEBN implements also a reasoner based on the PR-OWL/MEBN framework. Unfortunately, this GUI has a General Public License (GPL) which is a little too restrictive in order to integrate it into a commercial application. In addition, to our knowledge MEBN community is not wide enough to be considered as an emerging standard. Thus it represents a major difficulty to manipulate this tool.

### C. Fuzzy Approaches

Fuzzy formalisms allow the representation and the gradual assessment of truth about vague information.

FuzzyDL [10] is one of the most succeeded tools that propose a fuzzy description language associated to a reasoning engine. Even if the syntax is based on fuzzy SHIF, we can translate a basic ontology into a fuzzyDL file, thanks to a given specific API. Then fuzzy statements can be added to this file, or we can directly make use of the fuzzyOntology. The latter is an OWL ontology that may be used to encode Fuzzy OWL statements and thanks to a parser, it can be translated into a fuzzyDL file. Finally, it supports two types of reasoning: Zadeh semantics and Lukasiewicz Logic. However, we could point out that the implementation is only intended to be run either on MacOSX or Linux.

### D. Dempster-Shafer Approaches

Dempster–Shafer theory [11], also known as the theory of evidence or the theory of belief functions, was developed by Glenn Shafer in 1975 after the work made by Arthur P. Dempster. Beliefs in a hypothesis are calculated as the sum of the masses of all sets it encloses. It is often presented as a generalization of the probability theory. Briefly, it allows combining evidence from different sources and arriving at a degree of belief (represented by a belief function) that takes into account all the available evidence.

Regarding ontology, it has been yet most applied in areas such as inconsistency handling in OWL ontologies and ontology mapping ([14] and [15]). These areas start from the hypothesis that ontologies do not handle uncertainty, but produce uncertainty when grouping together a set of ontologies. However, one could notice that very recently a new approach [16] has been presented, which does not deal with ontology mapping or inconsistencies; it focuses rather on translating an OWL taxonomy into a directed evidential network based on the evidence theory. However, it deals only with classes and none open source tool is yet being available to test it.

## III. Modeling and Reasoning on Dempster-Shafer Theory

We are now going to introduce the obvious advantages of the theory of belief functions that lead us to seek a tool, which implements this theory. However, following this brief state of art, there is currently and to our knowledge no available tool based on Dempster-Shafer theory that meets our purposes that is to say to represent and reason about the uncertainty contained in our data/information.

### A. Advantages of Dempster-Shafer Theory

Inside the theory of uncertainty representation, we encountered an important difficulty when trying these tools in regards to the type of knowledge available in input of data fusion applications. As a matter of fact, the representation of information absence is badly taken into account by the theory of probability. Indeed, prior and conditional probabilities need to be specified into probabilistic methods. This requirement often lead in using a symmetry (minimax error) argument to assign prior probabilities to random variables (e.g. assigning 0.5 to binary values for which no information is available about which is more likely). However, any information contained in the missing priors and conditionals is not used in the Dempster–Shafer framework unless it can be obtained indirectly — and arguably is then available for calculation using Bayes equations.

One of the major advantages of Dempster–Shafer theory over probability theory is thus to allow one to specify a degree of ignorance in a situation instead of being forced to supply prior probabilities. This ability to explicitly model the degree of ignorance makes the theory very appealing.

Moreover, probabilistic approaches reason only on singletons. On the contrary, Dempster-Shafer theory enables us not only to affect belief on elementary hypotheses but also on composite ones. This last point illustrates the fact that Dempster-Shafer theory manages also imprecision and inaccuracies.

In addition, there is in the theory of Probabilities, a strong relation between an event and its negation, since its sum equals to 1. The evidence theory implies no relations between the existence or not of an event. Thus it models only the belief associated to a class, without influencing the belief allots to others classes.

Lastly the evidence combination rule of the Dempster-Shafer theory provides an interesting operator to integrate multiple pieces of information from different sources. Thus it is very helpful when working on pieces of information that come from various sources, as in Data Fusion applications. Finally, decision on the optimal hypothesis choice can be made in a flexible and rational manner.

To that point of view, the evidence theory is much more flexible than the probability theory. It permits to manage as well uncertainties as the inaccuracies and the ignorance.

### B. *Ontological Representation*

In order to represent the uncertainty contained in ontologies, it appears natural that it should be the instances that embody uncertainty. Indeed, in traditional ontology languages (as in object programming languages) the instances represent "real" world entities, which are a specific example of the structure defined by concepts or classes.

#### 1) *Use case*

The use case is placed at the first level of the JDL and it supposes that we have already built an ontology of a military battlefield. We want to instantiate this ontology with different pieces of information that come from various sensors and that capture some data/information directly from the battlefield of interest. However, frequently these sensors convey contradictory pieces of information. Moreover, these sensors can be given a certain degree of confidence and thus lead to the assessment of the mass value associated to each observation. Even if in evidence theory, the determination of mass functions is a crucial step, it is not our purpose here and this matter will not be handled in this paper.

For example, sensor A and B observe the same moving entity which moves towards a specific direction. Sensor A grants the confidence of 0.4 that this entity is a military tank with characteristics *i* and the one of 0.6 that it embodies either a tank with characteristics *i* or with characteristics *j*. But sensor B has observed with a 0.5 confidence that it is a tank with characteristics *i*. It grants 0.2 of its confidence to the fact that it is a military truck and 0.3 to its total ignorance.

In terms of instances, also called individuals or objects, sensor A and B would lead to the creation of two instances of the same concept *Military_Tank* that have, for example, the object property *hasCharacteristics* linked to different instances (related to characteristics *i* and *j*) and to the instantiation of concept *Military_Truck*.

According to the Dempster-Shafer formalism, Ω which is the universal set, that is to say, the set of all states under consideration will be defined in this case by Ω = { tank_i, tank_j, truck }. Hereafter is given the assignment of the belief masses according to sensor A and B to their observations:

- $m_A$({tank_i}) = 0.4 ; $m_A$({tank_i, tank_j}) = 0.6
- $m_B$({tank_i}) = 0.5 ; $m_B$({truck}) = 0.2 ; $m_B$({tank_i, tank_j, truck}) = 0.3

This use case is situated at the first level of the JDL model, but it could have also presented some situation where we handle uncertainty in higher level as presented in the introduction of this paper.

#### 2) *Model*

We modeled a Dempster-Shafer ontology that can be merged with any ontology and that enables us to instantiate its concepts in an uncertain manner.

This ontology is what we call an upper ontology, since one can use it in every area of knowledge, in other words it is non domain specific. It is a formal representation of the theory of Dempster-Shafer, as it proposes a shared understanding of the main concepts: mass, belief, plausibility, source, etc. The following figure illustrates the global structure of this ontology called DS-Ontology, in regards of the different classes, hierarchical relations (*has subclass*) and object/datatype properties (dotted arrows).

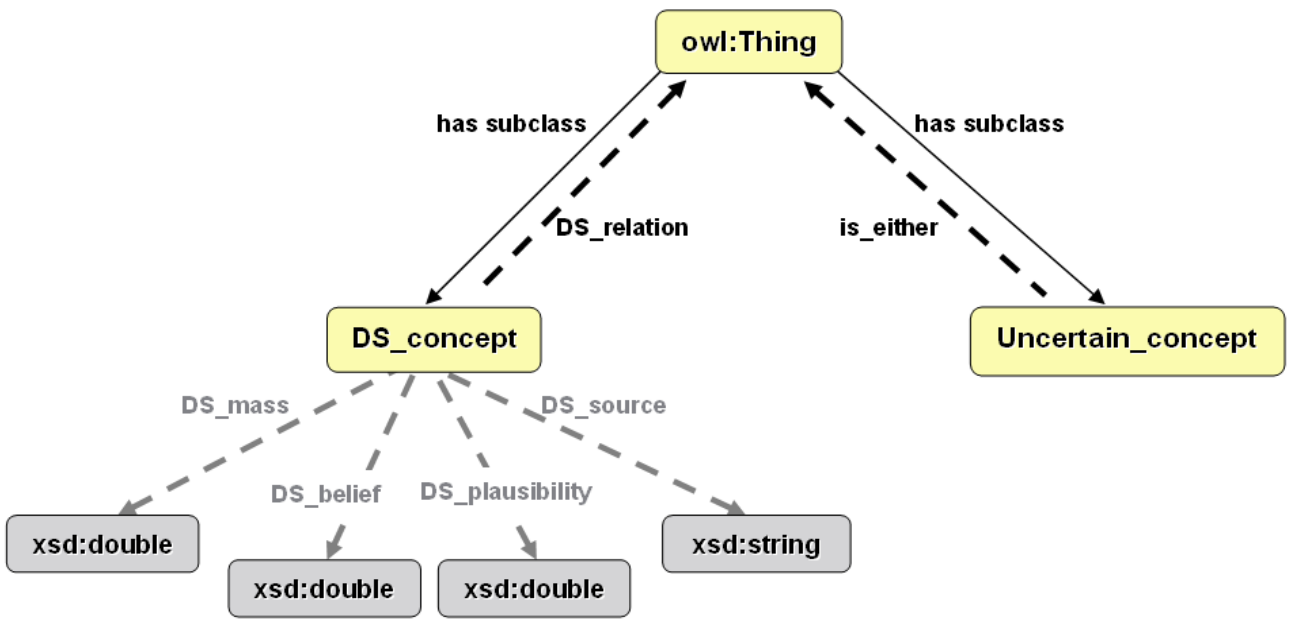

Figure 2. DS-Ontology structure

To import this DS-Ontology.owl file within a specific domain ontology, we use the *owl:imports* statement, such as: <owl:imports rdf:resource="DS-Ontology.owl"/>.

The *Uncertain_concept* is here to represent a concept that we want to instantiate in our domain ontology but it can refer to different classes, since sources convey different observations. This class (to be precise: its instance) gathers all the different individuals of our domain ontology it can embody thanks to the object property: *is_either*. One can note therefore that this property has for range the class: *owl:Thing* which is the root of all the classes, in order to accept all possible instances. For example, in the use case above, there would be only one instance of *Uncertain_concept*, linked to the three previously mentioned instances. The universal set whose definition is the set of all states under consideration is actually given by the list of instances linked to a specific instance of *Uncertain_concept*.

There is one *DS_concept* instance associated to each hypothesis, or composite hypotheses, thanks to the object property *DS_relation*. A *DS_concept* has datatype property: *DS_source*, which cardinality equals to 1. In other words, for each DS_concept, there is one DS_source associated. We proposed here to symbolize the source by a string (source "A", for instance). Then it also has the following properties: *DS_mass*, *DS_belief* and *DS_plausibility*, which have for value a double, which is implicitly ranging from 0 to 1. If we recall our initial use case for sensor A when dealing with the composite hypotheses {tank_i, tank_j}, the two instances of military tank will point to the same DS_concept object which has for *DS_source*: A and we assume for *DS_mass*: 0.6. A unique *DS_concept* will be linked with *DS_relation* to more or less instances according to the imprecision handled by the sensor.

Finally, we can finish filling in our domain ontology with instances that are not uncertain by the classical method, or when a certain instance is being involved in a relation that has for domain or range an uncertain concept, we bind it to the adequate instance of *Uncertain_concept*.

The following figure gives an outline of the instances that could be created regarding the initial use case.

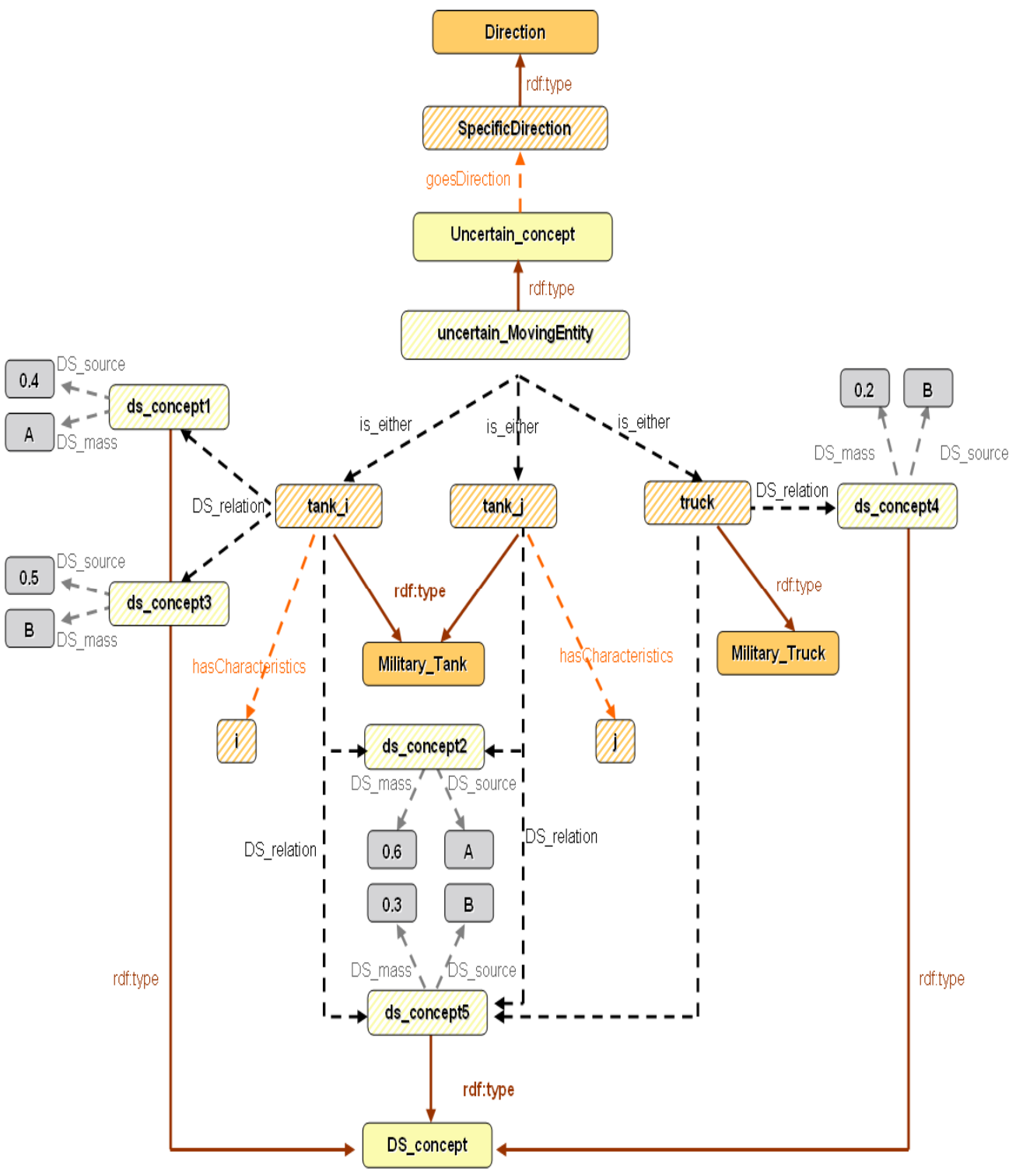

Figure 3. Instances example

To recap briefly: when we want to reason about an uncertain concept, we instantiate all the possible classes that this concept can embody and we link them with a specific instance of *Uncertain_concept* thanks to the relation: *is_either*. Then, we associate them to a *DS_concept* (thanks to *DS_relation*) and we can here declare which is the source that has made this observation and what is the value of mass, the belief or credibility that is associated to it.

## C. *Reasoning on uncertain ontological aspects*

Once having a unique general representation of Dempster-Shafer for any field, it has been quite easy to implement a Java application based on Jena framework that enables us to reason automatically about the uncertain objects contained in the ontology. Jena is an open source Java framework for building Semantic Web applications. It provides indeed a programmatic environment for RDF, RDFS and OWL, SPARQL and includes a rule-based inference engine. In our application, the Jena framework is a tool to retrieve the instances associated to each individual uncertain concept, and the Dempster-Shafer information collected. This information is then transmitted to a common basic Dempster-Shafer library [17] which then performs calculations in the Dempster-Shafer theory. In other words, our Java application first maps our ontology-based annotation (the set of related instances, for example those represented in Figure 3), written in OWL, into a Java representation of the Dempster-Shafer theory (according here to the Java library of [17]). Then this library enables us to manipulate the numerical values (mass, credibility, plausibility) which were previously contained in the annotation in order to perform classical calculations (conjunctive/disjunctive combination, conditioning, pignistic transformation, several uncertainty measures). As a matter of fact, our application calculates *K*, which represents the measure of the amount of conflict between mass sets. According to the value of K, we combine evidence from different sources, i.e. we combine the independent sets of mass assignments, thanks to Dempster's rule of combination. Finally, decision criteria can be applied to extract the chosen hypothesis, either based on the maximum of plausibility or belief function or either on pignistic probability criteria, etc. in regards to the type of problem handled.

As a consequence, the application creates a mapping between ontology representation and the Dempster-Shafer model which enables to reason about uncertainty in ontology.

## D. *Advantages, limits and future work*

For the moment, we have a simple mechanism to instantiate our domain ontology in an uncertain manner by importing the Dempster-Shafer ontology.

The advantage of this method is that one can come with any domain ontology and instantiate it, without making any change in this ontology. This is mostly due to the fact that OWL is a structured language but flexible enough to overcome the strict definition of properties which restricts the value of the range or domain to some given classes. What we mean is that without this flexibility, we would have to modify slightly the original ontology during the last phase of instantiation of certain classes in relation to instance of uncertain concept. Indeed, it would be impossible to link a property to an instance of *Uncertain_concept* (in the role of range or domain) since this class does not appear in the property definition in the domain ontology. Therefore we would have to add that the relation can accept for domain (or for range) an instance of *Uncertain_concept*. However, even if this is accepted by the syntax, it lowers naturally a little the expressivity of the ontology.

Future work can carry about the representation of uncertainty within relations (between concepts or between a concept and data value) in ontology.

Moreover, in a near future, we are also interested in using some reasoning capabilities of OWL language instead of simply use the ontology as a mean to represent the observations (observed entity and mass value, belief and/or plausibility associated to it) carried out by independent sources. As a matter of fact, OWL enables us, for example, to link resources which are instances of equivalent concepts defined under different labels ("owl:equivalentClass") or to reference a single individual using different names ("owl:sameAs"). Sources can thus use different ways to refer to identical resources. Therefore it would be conceivable to take them into account while applying the Dempster's rule of combination when these specific OWL properties are met in the domain ontology. In the same way, we could also enlarge this reasoning to the taxonomy of the ontology, by including the notion of semantic distance between two concepts. Indeed, two entities observed by different sources that are quite similar in nature (i.e. "closed" in the taxonomy) should have their belief augmented in comparison to a single entity observed, which is far from the others. Further work is needed in that field but a rule language (e.g. Jena rules [5], SWRL [18], RIF [19], etc.) could be used in order to perform those changes of mass values.

For other purposes, it could also be imagined that the application, which has an uncertain ontology for input, returns a non uncertain ontology, i.e. a traditional ontology, which would be the result of the Dempster's rule of combination, followed by a decision process on single hypothesis. However, for Data Fusion applications, it is preferable to keep this representation of uncertainty.

## IV. Conclusion

To conclude, let's remind that Data Fusion is the art of managing many pieces of information that are imperfect in multiple ways and to provide at the end a clearer view of the situation, that is to say less uncertain. Therefore there is currently a serious deficiency in ontological capacities. Thus, we have seen that uncertainty extension to OWL has started to make great strides in the last few years. We have proposed in this paper a simple approach to address the problem of representing uncertainty, based on the evidential approach. We have presented a mechanism to instantiate our domain ontology thanks to the DS-Ontology. The advantage of this method is that one can come with any domain ontology and instantiate it in an uncertain manner, without making any change in the domain ontology.

## Acknowledgment

We are grateful to Arnaud Saval and Khaled Khelif for their comments and hope that the revised paper has addressed their concerns.